\pdfoutput=1
\documentclass[11pt]{article}

\usepackage[]{EMNLP2023}

\usepackage{times}
\usepackage{latexsym}

\usepackage[T1]{fontenc}
\usepackage[utf8]{inputenc}

\usepackage{microtype}

\usepackage{inconsolata}

\usepackage{graphicx}
\usepackage{booktabs}

\title{IKUN for WMT24 General MT Task: \\ LLMs Are here for Multilingual Machine Translation}

\author{
  Baohao Liao$^{1,2}$ \:\:\: Christian Herold$^{2}$ \:\:\: Shahram Khadivi$^{2}$ \:\:\: Christof Monz$^{1}$ \\
  $^{1}$Language Technology Lab, University of Amsterdam \\
  $^{2}$eBay Inc., Aachen, Germany \\
  \texttt{\href{mailto:b.liao@uva.nl}{b.liao@uva.nl}}
}

\begin{document}
\maketitle
\begin{abstract}
This paper introduces two multilingual systems, \textit{IKUN} and \textit{IKUN-C}, developed for the general machine translation task in WMT24. IKUN and IKUN-C represent an \textit{open system} and a \textit{constrained system}, respectively, built on Llama-3-8b and Mistral-7B-v0.3. Both systems are designed to handle all 11 language directions using a single model. According to automatic evaluation metrics, \textbf{IKUN-C achieved 6 first-place and 3 second-place finishes among all constrained systems, while IKUN secured 1 first-place and 2 second-place finishes across both open and constrained systems}. These encouraging results suggest that large language models (LLMs) are nearing the level of proficiency required for effective multilingual machine translation. The systems are based on a two-stage approach: first, continuous pre-training on monolingual data in 10 languages, followed by fine-tuning on high-quality parallel data for 11 language directions. The primary difference between IKUN and IKUN-C lies in their monolingual pre-training strategy. IKUN-C is pre-trained using constrained monolingual data, whereas IKUN leverages monolingual data from the OSCAR dataset. In the second phase, both systems are fine-tuned on parallel data sourced from NTREX, Flores, and WMT16-23 for all 11 language pairs.
\end{abstract}

\section{Introduction}
Large language models (LLMs) \cite{llama2, llama3, mistral, DBLP:journals/corr/abs-2303-08774} serve as a crucial foundation for a wide range of applications. One significant advantage of LLMs is their ability to be applied across various tasks, thereby simplifying deployment processes. However, the application of LLMs to multilingual machine translation (MT) presents several challenges:

\begin{itemize}
    \item Most LLMs are pre-trained on one or a few dominant languages, making direct fine-tuning on multilingual data insufficient for ensuring optimal performance, particularly for low-resource languages, which are often underrepresented in the training data.
    \item It remains unclear whether these LLMs, primarily pre-trained on a limited number of languages, effectively facilitate transfer learning across different languages \cite{DBLP:journals/corr/abs-2404-11201}.
    \item The large-scale nature of most LLMs presents significant challenges for efficient fine-tuning, particularly for researchers and practitioners with limited computational resources.
\end{itemize}

In the WMT24 general MT task \cite{kocmi2024preliminarywmt24rankinggeneral}, our objective is to assess the capability of LLMs for multilingual MT, as an alternative to training bilingual systems from scratch \cite{DBLP:journals/corr/abs-2310-09946}. This paper provides a detailed account of how we developed our final multilingual system using LLMs for both the constrained and open tracks.

Firstly, we identified that certain LLMs exhibit inefficiencies in tokenizing sentences from languages that are underrepresented in the pre-training data. To address this, we extended the existing vocabulary to reduce the tokenized sentence length, thereby enhancing training efficiency. Secondly, we enriched the LLMs with knowledge across the 10 target languages through continued pre-training. This step is particularly crucial for underrepresented languages, as it facilitates transfer learning. Finally, we fine-tuned the models using high-quality parallel datasets across all 11 pairs.

Through this streamlined approach, our constrained multilingual system, IKUN-C, secured 6 first-place and 3 second-place rankings in the automatic evaluation. Our open multilingual system, IKUN, achieved 1 first-place and 2 second-place rankings across the open and constrained tracks. These encouraging results demonstrate that LLMs can be effectively adapted for multilingual MT, broadening access to speakers of diverse languages.

\begin{figure*}[t]
  \centering
  \includegraphics[width=0.98\linewidth]{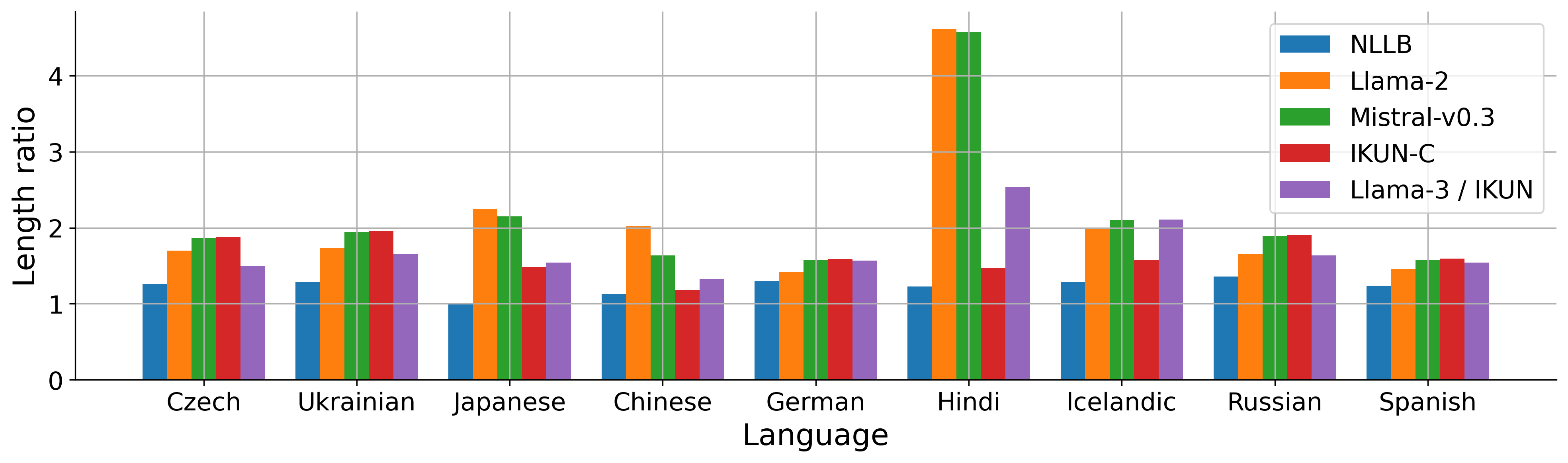}
  \caption{Tokenizer efficiency for various LLMs and languages. The larger the length ratio is, the less efficient the tokenizer is. We add new sub-words, from Chinese, Japanese, Hindi and Icelandic, to the Mistral-v0.3 vocabulary to construct the IKUN-C vocabulary. IKUN uses the Llama-3 tokenizer without any modification.}
  \label{fig: bpe length}
\end{figure*}
\section{Pre-trained LLM}
LLMs are pre-trained on extensive web-scale data, encompassing a vast repository of general knowledge applicable to various tasks. Previous studies \cite{alma, alma-r} have demonstrated that LLMs can substantially enhance the performance of multilingual MT. Building on this insight, we adopt a pre-trained LLM as the foundation for our system.

\textbf{IKUN} is an open system developed with meticulous consideration of available resources and system capabilities. For this purpose, we selected Llama-3 \cite{llama3}, one of the most advanced open-source LLMs available at the time of this competition. Due to constraints on computational resources, we opted for the 8B version\footnote{\href{https://huggingface.co/meta-llama/Meta-Llama-3-8B}{https://huggingface.co/meta-llama/Meta-Llama-3-8B}} instead of the 70B version. A significant factor in our choice of Llama-3 was its strong support for multilingual applications, as evidenced by the efficiency with which its tokenizer handles all 11 languages involved in this competition (See \S\ref{sec: tokenizer efficiency}). We also tried the instruct version, but it is worse than the pre-trained version.

\textbf{IKUN-C} is a constrained system based on Mistral-7B-v0.3 \cite{mistral}, one of the three LLMs permitted for the constrained track. Prior to selecting Mistral-7B-v0.3\footnote{\href{https://huggingface.co/mistralai/Mistral-7B-v0.3}{https://huggingface.co/mistralai/Mistral-7B-v0.3}}, we conducted continuous pre-training on all three allowed LLMs — namely, Llama-2-7B, Llama-2-13B \cite{llama2}, and Mistral-7B — using a subset of our monolingual data (approximately 1B tokens). The pre-training loss demonstrated that Mistral-7B outperformed Llama-2-7B and performed comparably to Llama-2-13B, leading us to select it as our architecture of choice.

\section{Tokenizer Efficiency}
\label{sec: tokenizer efficiency}
A significant challenge in applying LLMs to multilingual MT lies in the efficiency of their tokenizers. These models are typically pre-trained on one or a few dominant languages, and when their tokenizers are applied to low-resource languages, they produce disproportionately long sequences of sub-words. This inefficiency leads to excessive GPU memory consumption during training.

To evaluate the efficiency of the tokenizer, we focus on comparing the length differences between tokenized English sentences and their corresponding non-English counterparts. Specifically, we define the length ratio as:
\[
    \mathrm{length} \ \ \mathrm{ratio} = \frac{\mathrm{len(tokenizer(}x\mathrm{))}}{\mathrm{len(tokenizer(}y\mathrm{))}}
\]
where $y$ represents the English sentence, and $x$ denotes the paired non-English sentence. A smaller length ratio (close to 1) is desired, since it means that the tokenizer can encode the non-English sentence as efficient as the English sentence.

To facilitate a comparison of length ratios across different languages, English-centric multilingual data is essential. Fortunately, the FLoRes-200 dataset \cite{nllb} possesses this characteristic. In the devtest and test sets of FLoRes-200, every English sentence is paired with translations in various other languages. We concatenate all sentences from the devtest set and compute the length ratio for each language, as illustrated in Figure \ref{fig: bpe length}. We also include NLLB's tokenizer \cite{nllb} for a comparison.

We can observe: (1) NLLB consistently exhibits the smallest length ratio across all languages, likely due to its extensive optimization for hundreds of languages, thus serving as a lower bound in this context. (2) Mistral-v0.3 and Llama-3 demonstrate a notably high length ratio for Hindi, suggesting that Hindi is underrepresented in the pre-training data. (3) Compared to NLLB, the tokenizer of Mistral-v0.3 is significantly less efficient for Chinese, Japanese, Hindi, and Icelandic.

We opted to expand the vocabulary by incorporating new sub-words to reduce the length of tokenized sentences, thereby enhancing training efficiency. However, this approach introduces a trade-off between the addition of new sub-words and training performance. The embeddings for the newly introduced sub-words are initially untrained, and a substantial increase in sub-words may necessitate additional iterations of continuous pre-training. Consequently, our strategy for adding sub-words prioritizes those from languages with higher length ratios. 

For our open system, IKUN, we didn't modify its tokenizer, since Llama-3 tokenizer is already efficient enough, only except for Hindi. For our constrained system, IKUN-C, we expanded its vocabulary for Chinese, Japanese, Hindi, and Icelandic through the following steps: (1) Generate a new vocabulary of 12K sub-words using monolingual data from the past two years for these four languages from News Crawl\footnote{\href{https://data.statmt.org/news-crawl/}{https://data.statmt.org/news-crawl/}}; and (2) Merge the new vocabulary with the original. The efficiency of the resulted IKUN-C tokenizer is shown in Figure \ref{fig: bpe length}, demonstrating more efficiency for these four languages, especially for Hindi.

\begin{table}[t]
  \centering
  \small
  \begin{tabular}{crcr}
  \toprule
  \textbf{Language pair} & \textbf{Num.} & \textbf{Language pair} & \textbf{Num.} \\
  \midrule
  cs-uk & 8768 & uk-cs & 8768 \\
  ja-zh & 12858 & zh-ja & 12858 \\
  en-zh & 36647 & zh-en & 34650 \\
  en-cs & 30120 & cs-en & 28123 \\
  en-de & 35564 & de-en & 33567 \\
  en-hi & 11020 & hi-en & 9023 \\
  en-is & 8010 & is-en & 6013 \\
  en-ja & 18113 & ja-en & 16116 \\
  en-ru & 32840 & ru-en & 30843 \\
  en-es & 13808 & es-en & 11811 \\
  en-uk & 11961 & uk-en & 9964 \\
  en-fr & 4006 & fr-en & 4006 \\
  \midrule
  \textbf{Total:} & \multicolumn{3}{c}{429457} \\
  \bottomrule
  \end{tabular}
  \caption{Number of parallel sentences.\label{tab: parallel sentences}}
\end{table}

\begin{table*}[t]
  \centering
  \small
  \begin{tabular}{lll}
  \toprule
  \textbf{Hyper-parameter} & \textbf{Continuous pre-training} & \textbf{Finetuning} \\
  \midrule
  sampling probability & cs,de,en,es,hi,is,ja,ru,uk,zh = & \\
  & 0.1,0.13,0.1,0.13,0.08,0.05,0.08,0.13,0.08,0.12 & \\
  duration & 60K steps & 1 epoch \\
  batch size & 64 & 128 \\
  sequence length & 2048 & max source length=512, max target length=512\\
  learning rate (lr) & 2e-5 & 2e-4 \\
  warmup ratio & 0 & 0.01 \\
  weight decay & 0.01 & 0.01 \\
  lr scheduler & cosine & inverse\_sqrt \\
  training type & full finetuning & LoRA $r=64$ for all layers \\
  \bottomrule
  \end{tabular}
  \caption{Experimental setting for continuous pre-training and subsequent finetuning.\label{tab: setting}}
\end{table*}

\begin{table*}[t]
  \centering
  \scriptsize
  \begin{tabular}{llccccccccccc}
  \toprule
  \textbf{System} & \textbf{Metric} & \textbf{cs-uk} & \textbf{en-cs} & \textbf{en-de} & \textbf{en-es} & \textbf{en-hi} & \textbf{en-is} & \textbf{en-ja} & \textbf{en-ru} & \textbf{en-uk} & \textbf{en-zh} & \textbf{ja-zh} \\
  \midrule
  & MetricX $\downarrow$ & 2.4 & 4.3 & 2.0 & 3.5 & 7.1 & 4.9 & 4.3 & 4.7 & 4.7 & 4.2 & 6.2 \\
  & CometKiwi $\uparrow$ & 0.648 & 0.618 & 0.641 & 0.666 & 0.499 & 0.657 & 0.669 & 0.649 & 0.622 & 0.624 & 0.519 \\
  IKUN-C & AutoRank $\downarrow$ & 3.0 & 4.7 & 3.8 & 3.4 & 5.5 & 3.7 & 3.9 & 3.9 & 3.9 & 3.5 & 5.5 \\
  \cmidrule(lr){2-13}
  & Place in constrained $\downarrow$ & 2 & 5 & 1 & 1 & 1 & 1 & 3 & 1 & 1 & 2 & 2 \\
  \midrule
  & MetricX $\downarrow$ & 1.6 & 3.7 & 1.8 & 3.3 & 9.4 & 4.3 & 3.7 & 4.1 & 3.7 & 4.0 & 5.4 \\
  & CometKiwi $\uparrow$ & 0.664 & 0.638 & 0.668 & 0.687 & 0.428 & 0.666 & 0.696 & 0.675 & 0.661 & 0.646 & 0.544 \\
  IKUN & AutoRank $\downarrow$ & 2.3 & 3.9 & 3.0 & 2.8 & 7.7 & 3.2 & 3.1 & 3.2 & 2.8 & 3.1 & 4.4 \\
  \cmidrule(lr){2-13}
  & Place in constrained\&open $\downarrow$ & 2 & 5 & 4 & 3 & 5 & 1 & 6 & 3 & 2 & 5 & 6 \\
  \bottomrule
  \end{tabular}
  \caption{Preliminary results of our systems on the WMT24 test sets, taken from \citet{kocmi2024preliminarywmt24rankinggeneral}. The final human evaluation results are not released yet. ``-''  here means $\rightarrow$. I.e. cs-uk is cs$\rightarrow$uk. \label{tab: results}}
\end{table*}

\section{Experiments}
We mainly follow the pipeline from \citet{alma}, i.e. continuous pre-training on monolingual data for all 10 languages, and followed by fine-tuning on parallel data for all 11 language pairs.

\subsection{Continuous Pre-training}
Given that our selected LLMs, specifically Llama-3-8B and Mistral-7B-v0.3, are primarily pre-trained on English, it is necessary to incorporate knowledge from other languages through further pre-training, with particular emphasis on low-resource languages. Additionally, the word embeddings for the newly introduced sub-words in IKUN-C must also undergo training.

For the open system IKUN, we utilize monolingual data from the Oscar dataset \cite{oscar}, covering all 10 target languages. We adopt the sampling strategy outlined by \citet{alma}, described as:
\[
 P(l) \propto (\frac{D_l}{\sum_{l' \in L}D_{l'}})^{\frac{1}{T}} \quad s.t. \quad \sum_{l' \in L}P(l') = \frac{9}{10}
\]
where $D_l$ represents the number of words in language $l$\footnote{\href{https://huggingface.co/datasets/oscar-corpus/OSCAR-2301}{https://huggingface.co/datasets/oscar-corpus/OSCAR-2301}}, $T$ is the temperature parameter (set to 6), and $L$ denotes the set of all languages except English. The sampling probability for English is fixed at 1/10. The experimental settings for continuous pre-training are detailed in Table \ref{tab: setting}. We approximately pre-trained IKUN on an additional 8B tokens.

In the constrained system, only the provided data sources are permitted for use\footnote{\href{https://www2.statmt.org/wmt24/mtdata/}{https://www2.statmt.org/wmt24/mtdata/}}. The IKUN-C system utilizes monolingual data from the News Crawl dataset for 9 languages, with the exception of Spanish, as the use of Spanish monolingual data from News Crawl is restricted. For Spanish, we incorporate monolingual data from the Leipzig Corpora \cite{leipzig}. Additionally, for Hindi, we augment the dataset with monolingual data from News Commentary due to the relatively limited amount of Hindi data available in the News Crawl dataset. This adjustment is crucial because Hindi is underrepresented in the pre-training of Mistral-7B-v0.3. Our experimental settings closely align with those detailed in Table \ref{tab: setting}.

\subsection{Subsequent Fine-tuning}
Previous studies \cite{DBLP:journals/corr/abs-2401-12413, lima} have demonstrated that the quality of fine-tuning data is a critical factor in achieving optimal performance. \citet{DBLP:conf/wmt/LiaoKH21} further indicates that increasing the amount of back-translation data does not necessarily lead to better outcomes. In light of this, we exclusively utilize high-quality parallel data for the fine-tuning phase.

The high-quality parallel data is primarily sourced from FloRes-200 \cite{nllb}, NTREX-128 \cite{ntrex}, and previous WMT16-23 general MT/news tasks \cite{wmt23, wmt22, wmt21, wmt20, wmt19, wmt18, wmt17, wmt16}.

\textbf{FloRes-200:} As the FloRes-200 dataset provides parallel sentences across multiple languages, we leverage all 11 translation directions from the devtest and test sets. Importantly, our fine-tuning approach is not limited to the required directions listed in Table \ref{tab: results}; instead, we fine-tune the model on both translation directions, e.g., en$\leftrightarrow$de, to facilitate broader applicability of the final model.

\textbf{NTREX-128:} We also incorporate parallel sentences from NTREX-128 for from-English translation directions, i.e. en$\rightarrow$XX. In accordance with \citet{ntrex}, which recommends using the en$\rightarrow$XX translation direction, our fine-tuning is confined to these directions rather than adopting a bidirectional approach. An exception is made for the en-fr pair, where bidirectional fine-tuning is applied due to the limited availability of parallel data for this pair (absent in previous WMTs).

\textbf{Past WMTs:} Additionally, we extract parallel sentences from the development and test sets of the WMT16-23 general MT tasks, provided they contain the necessary translation directions. For these sentences, we employ a bidirectional fine-tuning strategy.

The statistics for all parallel sentences are presented in Table \ref{tab: parallel sentences}. Notably, all systems are fine-tuned at the sentence level. Given that WMT development and test sets are at the document level, models could alternatively be fine-tuned at the document level or reformatted into a conversational structure for fine-tuning. This latter approach might be more effective for context-aware translation, as WMT24 applies context-based human evaluations. We reserve this exploration for future work. The fine-tuning setting is listed in Table \ref{tab: setting}.

\subsection{Results}
We present the preliminary results reported by \citet{kocmi2024preliminarywmt24rankinggeneral} in Table \ref{tab: results}, which includes four evaluation metrics. Both MetricX-23-XL \cite{metricx} and CometKiwi-DA-XL \cite{cometkiwi} have demonstrated strong correlations with human evaluation \cite{DBLP:conf/wmt/FreitagMLARTKBD23}. AutoRank \cite{kocmi2024preliminarywmt24rankinggeneral}, a normalized composite metric derived from MetricX and CometKiwi, scales the scores of each metric linearly to span the range from 1 to the total number of systems in a given language pair. The final automatic ranking is obtained by averaging these normalized scores. AutoRank can thus be considered a measure of the overall rank across all systems and tracks. Additionally, we report the rankings of our systems across various tracks.

It is noteworthy that both of our systems are multilingual, designed to handle all language pairs. The IKUN-C system, in particular, demonstrates promising performance in the constrained track, achieving 6 first-place and 3 second-place finishes. In both the open and constrained tracks, IKUN maintains strong performance, securing 1 first-place and 2 second-place positions, even when compared to systems that may leverage additional open-source data or specialize in a limited set of language pairs.

\section{Conclusion}
In this paper, we present a methodology for effectively adapting pre-trained LLMs to the task of multilingual machine translation. Our approach involves three primary steps: (1) expanding the vocabulary to accommodate languages that are underrepresented in the pre-training data, when necessary; (2) continuing pre-training the LLM on monolingual data to enhance its knowledge of underrepresented languages and to train the embeddings of newly introduced sub-words; and (3) fine-tuning the LLM on high-quality parallel data. Our experimental results demonstrate the efficacy of this straightforward pipeline, with IKUN-C securing 6 first-place finishes in the constrained track, and IKUN achieving 1 first-place ranking in both the open and constrained tracks.

\section*{Acknowledgments}
We thank eBay Inc. for the computation support. This research was funded in part by the Netherlands Organization for Scientific Research (NWO) under project number VI.C.192.080.

\bibliography{custom}

\begin{thebibliography}{27}
\expandafter\ifx\csname natexlab\endcsname\relax\def\natexlab#1{#1}\fi

\bibitem[{Akhbardeh et~al.(2021)Akhbardeh, Arkhangorodsky, Biesialska, Bojar, Chatterjee, Chaudhary, Costa{-}juss{\`{a}}, Espa{\~{n}}a{-}Bonet, Fan, Federmann, Freitag, Graham, Grundkiewicz, Haddow, Harter, Heafield, Homan, Huck, Amponsah{-}Kaakyire, Kasai, Khashabi, Knight, Kocmi, Koehn, Lourie, Monz, Morishita, Nagata, Nagesh, Nakazawa, Negri, Pal, Tapo, Turchi, Vydrin, and Zampieri}]{wmt21}
Farhad Akhbardeh, Arkady Arkhangorodsky, Magdalena Biesialska, Ondrej Bojar, Rajen Chatterjee, Vishrav Chaudhary, Marta~R. Costa{-}juss{\`{a}}, Cristina Espa{\~{n}}a{-}Bonet, Angela Fan, Christian Federmann, Markus Freitag, Yvette Graham, Roman Grundkiewicz, Barry Haddow, Leonie Harter, Kenneth Heafield, Christopher Homan, Matthias Huck, Kwabena Amponsah{-}Kaakyire, Jungo Kasai, Daniel Khashabi, Kevin Knight, Tom Kocmi, Philipp Koehn, Nicholas Lourie, Christof Monz, Makoto Morishita, Masaaki Nagata, Ajay Nagesh, Toshiaki Nakazawa, Matteo Negri, Santanu Pal, Allahsera~Auguste Tapo, Marco Turchi, Valentin Vydrin, and Marcos Zampieri. 2021.
\newblock \href {https://aclanthology.org/2021.wmt-1.1} {Findings of the 2021 conference on machine translation {(WMT21)}}.
\newblock In \emph{Proceedings of the Sixth Conference on Machine Translation, WMT@EMNLP 2021, Online Event, November 10-11, 2021}, pages 1--88. Association for Computational Linguistics.

\bibitem[{Barrault et~al.(2020)Barrault, Biesialska, Bojar, Costa{-}juss{\`{a}}, Federmann, Graham, Grundkiewicz, Haddow, Huck, Joanis, Kocmi, Koehn, Lo, Ljubesic, Monz, Morishita, Nagata, Nakazawa, Pal, Post, and Zampieri}]{wmt20}
Lo{\"{\i}}c Barrault, Magdalena Biesialska, Ondrej Bojar, Marta~R. Costa{-}juss{\`{a}}, Christian Federmann, Yvette Graham, Roman Grundkiewicz, Barry Haddow, Matthias Huck, Eric Joanis, Tom Kocmi, Philipp Koehn, Chi{-}kiu Lo, Nikola Ljubesic, Christof Monz, Makoto Morishita, Masaaki Nagata, Toshiaki Nakazawa, Santanu Pal, Matt Post, and Marcos Zampieri. 2020.
\newblock \href {https://aclanthology.org/2020.wmt-1.1/} {Findings of the 2020 conference on machine translation {(WMT20)}}.
\newblock In \emph{Proceedings of the Fifth Conference on Machine Translation, WMT@EMNLP 2020, Online, November 19-20, 2020}, pages 1--55. Association for Computational Linguistics.

\bibitem[{Barrault et~al.(2019)Barrault, Bojar, Costa{-}juss{\`{a}}, Federmann, Fishel, Graham, Haddow, Huck, Koehn, Malmasi, Monz, M{\"{u}}ller, Pal, Post, and Zampieri}]{wmt19}
Lo{\"{\i}}c Barrault, Ondrej Bojar, Marta~R. Costa{-}juss{\`{a}}, Christian Federmann, Mark Fishel, Yvette Graham, Barry Haddow, Matthias Huck, Philipp Koehn, Shervin Malmasi, Christof Monz, Mathias M{\"{u}}ller, Santanu Pal, Matt Post, and Marcos Zampieri. 2019.
\newblock \href {https://doi.org/10.18653/V1/W19-5301} {Findings of the 2019 conference on machine translation {(WMT19)}}.
\newblock In \emph{Proceedings of the Fourth Conference on Machine Translation, {WMT} 2019, Florence, Italy, August 1-2, 2019 - Volume 2: Shared Task Papers, Day 1}, pages 1--61. Association for Computational Linguistics.

\bibitem[{Barrault et~al.(2018)Barrault, Bougares, Specia, Lala, Elliott, and Frank}]{wmt18}
Lo{\"{\i}}c Barrault, Fethi Bougares, Lucia Specia, Chiraag Lala, Desmond Elliott, and Stella Frank. 2018.
\newblock \href {https://doi.org/10.18653/V1/W18-6402} {Findings of the third shared task on multimodal machine translation}.
\newblock In \emph{Proceedings of the Third Conference on Machine Translation: Shared Task Papers, {WMT} 2018, Belgium, Brussels, October 31 - November 1, 2018}, pages 304--323. Association for Computational Linguistics.

\bibitem[{Bojar et~al.(2017)Bojar, Chatterjee, Federmann, Graham, Haddow, Huang, Huck, Koehn, Liu, Logacheva, Monz, Negri, Post, Rubino, Specia, and Turchi}]{wmt17}
Ondrej Bojar, Rajen Chatterjee, Christian Federmann, Yvette Graham, Barry Haddow, Shujian Huang, Matthias Huck, Philipp Koehn, Qun Liu, Varvara Logacheva, Christof Monz, Matteo Negri, Matt Post, Raphael Rubino, Lucia Specia, and Marco Turchi. 2017.
\newblock \href {https://doi.org/10.18653/V1/W17-4717} {Findings of the 2017 conference on machine translation {(WMT17)}}.
\newblock In \emph{Proceedings of the Second Conference on Machine Translation, {WMT} 2017, Copenhagen, Denmark, September 7-8, 2017}, pages 169--214. Association for Computational Linguistics.

\bibitem[{Bojar et~al.(2016)Bojar, Chatterjee, Federmann, Graham, Haddow, Huck, Jimeno{-}Yepes, Koehn, Logacheva, Monz, Negri, N{\'{e}}v{\'{e}}ol, Neves, Popel, Post, Rubino, Scarton, Specia, Turchi, Verspoor, and Zampieri}]{wmt16}
Ondrej Bojar, Rajen Chatterjee, Christian Federmann, Yvette Graham, Barry Haddow, Matthias Huck, Antonio Jimeno{-}Yepes, Philipp Koehn, Varvara Logacheva, Christof Monz, Matteo Negri, Aur{\'{e}}lie N{\'{e}}v{\'{e}}ol, Mariana~L. Neves, Martin Popel, Matt Post, Raphael Rubino, Carolina Scarton, Lucia Specia, Marco Turchi, Karin Verspoor, and Marcos Zampieri. 2016.
\newblock \href {https://doi.org/10.18653/V1/W16-2301} {Findings of the 2016 conference on machine translation}.
\newblock In \emph{Proceedings of the First Conference on Machine Translation, {WMT} 2016, colocated with {ACL} 2016, August 11-12, Berlin, Germany}, pages 131--198. The Association for Computer Linguistics.

\bibitem[{Costa{-}juss{\`{a}} et~al.(2022)Costa{-}juss{\`{a}}, Cross, {\c{C}}elebi, Elbayad, Heafield, and et~al.}]{nllb}
Marta~R. Costa{-}juss{\`{a}}, James Cross, Onur {\c{C}}elebi, Maha Elbayad, Kenneth Heafield, and et~al. 2022.
\newblock \href {https://doi.org/10.48550/ARXIV.2207.04672} {No language left behind: Scaling human-centered machine translation}.
\newblock \emph{CoRR}, abs/2207.04672.

\bibitem[{Dubey et~al.(2024)Dubey, Jauhri, Pandey, Kadian, Al-Dahle, and et~al.}]{llama3}
Abhimanyu Dubey, Abhinav Jauhri, Abhinav Pandey, Abhishek Kadian, Ahmad Al-Dahle, and et~al. 2024.
\newblock \href {http://arxiv.org/abs/2407.21783} {The llama 3 herd of models}.

\bibitem[{Federmann et~al.(2022)Federmann, Kocmi, and Xin}]{ntrex}
Christian Federmann, Tom Kocmi, and Ying Xin. 2022.
\newblock \href {https://aclanthology.org/2022.sumeval-1.4} {{NTREX}-128 {--} news test references for {MT} evaluation of 128 languages}.
\newblock In \emph{Proceedings of the First Workshop on Scaling Up Multilingual Evaluation}, pages 21--24, Online. Association for Computational Linguistics.

\bibitem[{Freitag et~al.(2023)Freitag, Mathur, Lo, Avramidis, Rei, Thompson, Kocmi, Blain, Deutsch, Stewart, Zerva, Castilho, Lavie, and Foster}]{DBLP:conf/wmt/FreitagMLARTKBD23}
Markus Freitag, Nitika Mathur, Chi{-}kiu Lo, Eleftherios Avramidis, Ricardo Rei, Brian Thompson, Tom Kocmi, Fr{\'{e}}d{\'{e}}ric Blain, Daniel Deutsch, Craig Stewart, Chrysoula Zerva, Sheila Castilho, Alon Lavie, and George~F. Foster. 2023.
\newblock \href {https://doi.org/10.18653/V1/2023.WMT-1.51} {Results of {WMT23} metrics shared task: Metrics might be guilty but references are not innocent}.
\newblock In \emph{Proceedings of the Eighth Conference on Machine Translation, {WMT} 2023, Singapore, December 6-7, 2023}, pages 578--628. Association for Computational Linguistics.

\bibitem[{Goldhahn et~al.(2012)Goldhahn, Eckart, and Quasthoff}]{leipzig}
Dirk Goldhahn, Thomas Eckart, and Uwe Quasthoff. 2012.
\newblock \href {http://www.lrec-conf.org/proceedings/lrec2012/summaries/327.html} {Building large monolingual dictionaries at the leipzig corpora collection: From 100 to 200 languages}.
\newblock In \emph{Proceedings of the Eighth International Conference on Language Resources and Evaluation, {LREC} 2012, Istanbul, Turkey, May 23-25, 2012}, pages 759--765. European Language Resources Association {(ELRA)}.

\bibitem[{Jiang et~al.(2023)Jiang, Sablayrolles, Mensch, Bamford, Chaplot, and et~al.}]{mistral}
Albert~Q. Jiang, Alexandre Sablayrolles, Arthur Mensch, Chris Bamford, Devendra~Singh Chaplot, and et~al. 2023.
\newblock \href {https://doi.org/10.48550/ARXIV.2310.06825} {Mistral 7b}.
\newblock \emph{CoRR}, abs/2310.06825.

\bibitem[{Juraska et~al.(2023)Juraska, Finkelstein, Deutsch, Siddhant, Mirzazadeh, and Freitag}]{metricx}
Juraj Juraska, Mara Finkelstein, Daniel Deutsch, Aditya Siddhant, Mehdi Mirzazadeh, and Markus Freitag. 2023.
\newblock \href {https://doi.org/10.18653/v1/2023.wmt-1.63} {{M}etric{X}-23: The {G}oogle submission to the {WMT} 2023 metrics shared task}.
\newblock In \emph{Proceedings of the Eighth Conference on Machine Translation}, pages 756--767, Singapore. Association for Computational Linguistics.

\bibitem[{Kocmi et~al.(2024)Kocmi, Avramidis, Bawden, Bojar, Dvorkovich, Federmann, Fishel, Freitag, Gowda, Grundkiewicz, Haddow, Karpinska, Koehn, Marie, Murray, Nagata, Popel, Popovic, Shmatova, Steingrímsson, and Zouhar}]{kocmi2024preliminarywmt24rankinggeneral}
Tom Kocmi, Eleftherios Avramidis, Rachel Bawden, Ondrej Bojar, Anton Dvorkovich, Christian Federmann, Mark Fishel, Markus Freitag, Thamme Gowda, Roman Grundkiewicz, Barry Haddow, Marzena Karpinska, Philipp Koehn, Benjamin Marie, Kenton Murray, Masaaki Nagata, Martin Popel, Maja Popovic, Mariya Shmatova, Steinþór Steingrímsson, and Vilém Zouhar. 2024.
\newblock \href {http://arxiv.org/abs/2407.19884} {Preliminary wmt24 ranking of general mt systems and llms}.

\bibitem[{Kocmi et~al.(2023)Kocmi, Avramidis, Bawden, Bojar, Dvorkovich, Federmann, Fishel, Freitag, Gowda, Grundkiewicz, Haddow, Koehn, Marie, Monz, Morishita, Murray, Nagata, Nakazawa, Popel, Popovic, and Shmatova}]{wmt23}
Tom Kocmi, Eleftherios Avramidis, Rachel Bawden, Ondrej Bojar, Anton Dvorkovich, Christian Federmann, Mark Fishel, Markus Freitag, Thamme Gowda, Roman Grundkiewicz, Barry Haddow, Philipp Koehn, Benjamin Marie, Christof Monz, Makoto Morishita, Kenton Murray, Makoto Nagata, Toshiaki Nakazawa, Martin Popel, Maja Popovic, and Mariya Shmatova. 2023.
\newblock \href {https://doi.org/10.18653/V1/2023.WMT-1.1} {Findings of the 2023 conference on machine translation {(WMT23):} llms are here but not quite there yet}.
\newblock In \emph{Proceedings of the Eighth Conference on Machine Translation, {WMT} 2023, Singapore, December 6-7, 2023}, pages 1--42. Association for Computational Linguistics.

\bibitem[{Kocmi et~al.(2022)Kocmi, Bawden, Bojar, Dvorkovich, Federmann, Fishel, Gowda, Graham, Grundkiewicz, Haddow, Knowles, Koehn, Monz, Morishita, Nagata, Nakazawa, Nov{\'{a}}k, Popel, and Popovic}]{wmt22}
Tom Kocmi, Rachel Bawden, Ondrej Bojar, Anton Dvorkovich, Christian Federmann, Mark Fishel, Thamme Gowda, Yvette Graham, Roman Grundkiewicz, Barry Haddow, Rebecca Knowles, Philipp Koehn, Christof Monz, Makoto Morishita, Masaaki Nagata, Toshiaki Nakazawa, Michal Nov{\'{a}}k, Martin Popel, and Maja Popovic. 2022.
\newblock \href {https://aclanthology.org/2022.wmt-1.1} {Findings of the 2022 conference on machine translation {(WMT22)}}.
\newblock In \emph{Proceedings of the Seventh Conference on Machine Translation, {WMT} 2022, Abu Dhabi, United Arab Emirates (Hybrid), December 7-8, 2022}, pages 1--45. Association for Computational Linguistics.

\bibitem[{Liao et~al.(2021)Liao, Khadivi, and Hewavitharana}]{DBLP:conf/wmt/LiaoKH21}
Baohao Liao, Shahram Khadivi, and Sanjika Hewavitharana. 2021.
\newblock \href {https://aclanthology.org/2021.wmt-1.50} {Back-translation for large-scale multilingual machine translation}.
\newblock In \emph{Proceedings of the Sixth Conference on Machine Translation, WMT@EMNLP 2021, Online Event, November 10-11, 2021}, pages 418--424. Association for Computational Linguistics.

\bibitem[{OpenAI(2023)}]{DBLP:journals/corr/abs-2303-08774}
OpenAI. 2023.
\newblock \href {https://doi.org/10.48550/ARXIV.2303.08774} {{GPT-4} technical report}.
\newblock \emph{CoRR}, abs/2303.08774.

\bibitem[{Rei et~al.(2023)Rei, Guerreiro, Pombal, van Stigt, Treviso, Coheur, de~Souza, and Martins}]{cometkiwi}
Ricardo Rei, Nuno~Miguel Guerreiro, Jos{\'{e}} Pombal, Daan van Stigt, Marcos~V. Treviso, Lu{\'{\i}}sa Coheur, Jos{\'{e}} G.~C. de~Souza, and Andr{\'{e}} F.~T. Martins. 2023.
\newblock \href {https://doi.org/10.18653/V1/2023.WMT-1.73} {Scaling up cometkiwi: Unbabel-ist 2023 submission for the quality estimation shared task}.
\newblock In \emph{Proceedings of the Eighth Conference on Machine Translation, {WMT} 2023, Singapore, December 6-7, 2023}, pages 841--848. Association for Computational Linguistics.

\bibitem[{Su{\'{a}}rez et~al.(2020)Su{\'{a}}rez, Romary, and Sagot}]{oscar}
Pedro Javier~Ortiz Su{\'{a}}rez, Laurent Romary, and Beno{\^{\i}}t Sagot. 2020.
\newblock \href {https://doi.org/10.18653/V1/2020.ACL-MAIN.156} {A monolingual approach to contextualized word embeddings for mid-resource languages}.
\newblock In \emph{Proceedings of the 58th Annual Meeting of the Association for Computational Linguistics, {ACL} 2020, Online, July 5-10, 2020}, pages 1703--1714. Association for Computational Linguistics.

\bibitem[{Tan et~al.(2024)Tan, Wu, and Monz}]{DBLP:journals/corr/abs-2404-11201}
Shaomu Tan, Di~Wu, and Christof Monz. 2024.
\newblock \href {https://doi.org/10.48550/ARXIV.2404.11201} {Neuron specialization: Leveraging intrinsic task modularity for multilingual machine translation}.
\newblock \emph{CoRR}, abs/2404.11201.

\bibitem[{Touvron et~al.(2023)Touvron, Martin, Stone, Albert, Almahairi, and et~al.}]{llama2}
Hugo Touvron, Louis Martin, Kevin Stone, Peter Albert, Amjad Almahairi, and et~al. 2023.
\newblock \href {https://doi.org/10.48550/ARXIV.2307.09288} {Llama 2: Open foundation and fine-tuned chat models}.
\newblock \emph{CoRR}, abs/2307.09288.

\bibitem[{Wu et~al.(2024)Wu, Tan, Meng, Stap, and Monz}]{DBLP:journals/corr/abs-2401-12413}
Di~Wu, Shaomu Tan, Yan Meng, David Stap, and Christof Monz. 2024.
\newblock \href {https://doi.org/10.48550/ARXIV.2401.12413} {How far can 100 samples go? unlocking overall zero-shot multilingual translation via tiny multi-parallel data}.
\newblock \emph{CoRR}, abs/2401.12413.

\bibitem[{Wu et~al.(2023)Wu, Tan, Stap, Araabi, and Monz}]{DBLP:journals/corr/abs-2310-09946}
Di~Wu, Shaomu Tan, David Stap, Ali Araabi, and Christof Monz. 2023.
\newblock \href {https://doi.org/10.48550/ARXIV.2310.09946} {Uva-mt's participation in the {WMT23} general translation shared task}.
\newblock \emph{CoRR}, abs/2310.09946.

\bibitem[{Xu et~al.(2024{\natexlab{a}})Xu, Kim, Sharaf, and Awadalla}]{alma}
Haoran Xu, Young~Jin Kim, Amr Sharaf, and Hany~Hassan Awadalla. 2024{\natexlab{a}}.
\newblock \href {https://openreview.net/forum?id=farT6XXntP} {A paradigm shift in machine translation: Boosting translation performance of large language models}.
\newblock In \emph{The Twelfth International Conference on Learning Representations, {ICLR} 2024, Vienna, Austria, May 7-11, 2024}. OpenReview.net.

\bibitem[{Xu et~al.(2024{\natexlab{b}})Xu, Sharaf, Chen, Tan, Shen, Durme, Murray, and Kim}]{alma-r}
Haoran Xu, Amr Sharaf, Yunmo Chen, Weiting Tan, Lingfeng Shen, Benjamin~Van Durme, Kenton Murray, and Young~Jin Kim. 2024{\natexlab{b}}.
\newblock \href {https://doi.org/10.48550/ARXIV.2401.08417} {Contrastive preference optimization: Pushing the boundaries of {LLM} performance in machine translation}.
\newblock \emph{CoRR}, abs/2401.08417.

\bibitem[{Zhou et~al.(2023)Zhou, Liu, Xu, Iyer, Sun, Mao, Ma, Efrat, Yu, Yu, Zhang, Ghosh, Lewis, Zettlemoyer, and Levy}]{lima}
Chunting Zhou, Pengfei Liu, Puxin Xu, Srinivasan Iyer, Jiao Sun, Yuning Mao, Xuezhe Ma, Avia Efrat, Ping Yu, Lili Yu, Susan Zhang, Gargi Ghosh, Mike Lewis, Luke Zettlemoyer, and Omer Levy. 2023.
\newblock \href {http://papers.nips.cc/paper\_files/paper/2023/hash/ac662d74829e4407ce1d126477f4a03a-Abstract-Conference.html} {{LIMA:} less is more for alignment}.
\newblock In \emph{Advances in Neural Information Processing Systems 36: Annual Conference on Neural Information Processing Systems 2023, NeurIPS 2023, New Orleans, LA, USA, December 10 - 16, 2023}.

\end{thebibliography}
\bibliographystyle{acl_natbib}

\appendix
\end{document}